\documentclass{article}

\PassOptionsToPackage{numbers, compress}{natbib}
\usepackage[final]{nips_2017}
\usepackage[utf8]{inputenc} 
\usepackage[T1]{fontenc}    
\usepackage{hyperref}       
\usepackage{url}            
\usepackage{booktabs}       
\usepackage{amsfonts}       
\usepackage{nicefrac}       
\usepackage{microtype}      

\usepackage{amssymb,amsmath,array}
\usepackage{algorithm,algorithmic,mathtools}

\DeclareMathOperator*{\argmin}{argmin}

\title{Subject Selection on a Riemannian Manifold for Unsupervised Cross-subject Seizure Detection}

\author{
	Samaneh ~Nasiri Ghosheh Bolagh$^{(\ast)}$ , Gari. D. ~ Clifford$^{(\ast, \dagger)}$ \\
	Department of Biomedical Informatics, Emory University School of Medicine $^{(\ast)}$ \\ 
	Department of Biomedical Engineering, Georgia Institute of Technology$^{(\dagger)}$ \\
	\texttt{snasiri@emory.edu} \\
	\texttt{gari@gatech.edu} \\}

\begin{document}
	
	\maketitle
	
	\begin{abstract}
		Inter-subject variability between individuals poses a challenge in inter-subject brain signal analysis problems. 
		A new algorithm for subject-selection based on clustering covariance matrices on a Riemannian manifold is proposed. 
		After unsupervised selection of the subsets of relevant subjects, data in a cluster is mapped to a tangent space at the mean point of covariance matrices in that cluster and an SVM classifier on labeled data from relevant subjects is trained. 
		Experiment on an EEG seizure database shows that the proposed method increases the accuracy over state-of-the-art from 86.83\% to 89.84\% and specificity from  87.38\% to 89.64\% while reducing the false positive rate/hour from 0.8/hour to 0.77/hour.
	\end{abstract}
	\section{Introduction}
	Brain signal classification plays a crucial role in understanding underlying mental processes \cite{EEGClassification}. Perhaps the simplest and most convenient modality for analyzing the brain is through the electroencephalogram (EEG).  Classification of EEG features can be used for voluntary and involuntary tasks such as controlling the neuroprosthesis and detection of epileptic seizures \cite{EEGClassification}. It is well-known that brain signals are subject-specific, thus brain decoding models are traditionally designed individually: training and test data belongs to the same subject \cite{ESANN, TransferLearningRiemannian}.	However, it is often inconvenient, expensive and time-consuming to obtain a large number of training samples to train an associated classifier for every subject. To mitigate these issues, subsets of past subjects can be used to initialize a classifier for training on a new subject \cite{TransferLearningRiemannian}. This form of learning is referred to as cross-subject learning \cite{TransferLearningRiemannian}.
	However, this method is less effective, due to inter-subject variability and possibly large differences between the current subject and past training data.
	To address this problem, we use a clustering method on Riemannian manifold, which groups subjects based on the similarity of the features used by the classifier. 
	\par Barachant {\it et al.} \cite{multi} proposed a simple classifier based on the Riemannian distance for brain-computer interface classification. Appropriate forms of data covariance matrices were used as features and classification was performed based on a simple distance metric \cite{multi}. This approach provided superior cross-subject generalization capabilities compared to earlier works, as well as robustness to EEG artifacts, outliers and mislabeling \cite{ESANN,TransferLearningRiemannian}.
	
	In this article, we focus on the problem of cross-subject brain signal classification without assuming knowledge of labels from the new (test) subject. The key contribution of this work is to develop a subject selection algorithm on a Riemannian manifold. Clearly, the inherent inter-subject variability between individuals poses a challenge for the brain signal classification, therefore it is important to identify subsets of subjects that most closely map to the test case, and allow for a selection of a model trained completely or weighted towards such a subset. In order to obtain subsets of relevant subjects, we apply spectral clustering on a Riemannian manifold. Subsequently, a support vector machine (SVM) classifier is used in the tangent space to classify signals. Finally, the algorithm's performance is assessed on the open access CHB-MIT database \cite{physionet}. %
	\section{Subject Selection on a Riemannian Manifold}
	In machine learning, it is often assumed that the training and test subjects are drawn from the same distribution. Moreover, the patients on which we train the data should exhibit enough variation to allow an accurate sampling of the distribution. However, in the cross-subject classification problem, the distribution of training subjects and test subject are different from each other due to inherently high inter-subject variability and our inability to accurately sample the feature space for the patient population. Consequently, the training data comprises several sub-distributions. Therefore, identifying (and using) the subset of training subjects that are most similar to our test subject is likely to improve classification. To find the similar subjects, spectral clustering is applied on the feature space.	
	
	To describe the extraction of a feature, we let $\small \mathbf{X}_i \in \Re^{C \times T_s}$ denote a trial indexed by $i$, with $C$ the number of channels, $T_s$ the number of time samples and $y_i$ the class label of the trial. The feature covariance matrices are obtained simply by using a Sample Covariance Matrix (SCM) estimator \cite{kernelriem}, such as 
	\begin{equation}
	\small
	\label{eq:SCM}
	\mathbf{\Sigma}_i  = \frac{1}{T_s -1} {\mathbf{X}}_i {\mathbf{X}}_i^T.
	\end{equation}
	Such second-order information has been shown to be suitable for capturing the relevant information about mental states \cite{TransferLearningRiemannian}. The covariance matrices in (\ref{eq:SCM}) belong to the space of Symmetric Positive Definite (SPD) matrice, which form a Riemannian manifold  \cite{kernelriem}. Therefore, we can use tools  from differential geometry on Riemannian manifolds to manipulate the features. 
	\par \textbf{Riemannian Distance}:
	For any two covariance matrices $ \small \mathbf{\Sigma}_1$ and $\small \mathbf{\Sigma}_2 $, the Riemannian distance is defined according to the Riemannian metric as \cite{multi}
	\begin{equation}
	\small
	\label{eq:Rgeodistance}
	\delta_R(\mathbf{\Sigma}_1,\mathbf{\Sigma}_2) 
	= 
	\Vert \mathrm{log} \left( \mathbf{\Sigma}_1^{-1/2} \mathbf{\Sigma}_2 \mathbf{\Sigma}_1^{-1/2} \right) \Vert_F
	=
	\left[ \sum_{c=1}^{C} \log^2 \lambda_c \right]^{1/2}
	\end{equation}
	where $\lambda_c, c=1\ldots C$ are the real eigenvalues 
	of $\small \mathbf{\Sigma}_1^{-1/2} \mathbf{\Sigma}_2 \mathbf{\Sigma}_1^{-1/2}$ and $C$ the number of channels.
	This distance is \emph{Affine-invariant} \cite{multi}, i.e, it is invariant with respect to similar and congruent transformations, and inversion.
	\par \textbf{Riemannian Mean}:
	The Riemannian geometric mean of $I$ covariance matrices, 
	also called the Fr\'echet or Karcher mean, is the point on the manifold minimizing the dispersion given by \cite{multi}:
	\begin{equation}
	\small
	\label{eq:mean}
	\mathfrak{G} \left( \mathbf{\Sigma}_1,\ldots,\mathbf{\Sigma}_I \right) = \argmin_{\mathbf{\Sigma}} 
	\sum_{i=1}^{I} 
	\delta_R^2 \left( \mathbf{\Sigma},\mathbf{\Sigma}_i \right).
	\end{equation}
	There is no closed form expression for the mean of $I>2$. However a gradient descent procedure in the manifold can be used in order to find the solution \cite{kernelriem}. 
	
	Zianai et al \cite{TransferLearningRiemannian} have proposed to transform the covariance matrices of every subject in order to center them with respect to a reference covarince matrix, making the data from different subjects comparable. We use this transformation in our work., which is given by 
	\begin{equation}
	\label{eq:Transformation}
	\Sigma_i ^{(j)} \Rightarrow (\bar{\Sigma}^{(j)})^{-\frac{1}{2}} \Sigma_i ^{(j)} (\bar{\Sigma}^{(j)})^{-\frac{1}{2}}
	\end{equation}
	where $\bar{\Sigma}^{(j)}$ is the center of mass of covarinace matrices for subject $j$. 
	In order to find relevant subjects, spectral clustering is applied on Riemannian manifold. At first, an affinity matrix is defined using Gaussian function and the Riemannian distance. Following \cite{ng2002spectral}, the diagonal elements of the affinity matrix are set to zero, giving
	\begin{equation}
	\label{eq:Affinity}
	A_{ij}=exp(-\frac{\delta_R^2 (\bar{\mathbf{\Sigma}_i},\bar{\mathbf{\Sigma}_j})}{\sigma^2})
	\end{equation}
	The width of RBF kernel was chosen as $\sigma=0.5$, without attempting to optimize, since the database is not large enough and we are likely to overfit. 
	\par 
	Based on \cite{SpectralClusteringforTimeSeries}, the number of clusters can be estimated by the number of eigenvalues of $\mathbf{A}'$, the normalized affinity matrix, which are approximately equal to unity. In this work, based on the computed affinity matrix, the number of clusters was found to be five. 
	
	Algorithm 1 summarizes the proposed method for subject selection using spectral clustering on a Riemannian manifold. 
    
\begin{algorithm}
\scriptsize
  \caption{\small Subject Selection on Riemannian Manifold}
  \begin{algorithmic}[1]\label{alg:algo1}
    \REQUIRE $\mathbf{X}_t^{u}$  Unlabeled Signal from all subjects
   \ENSURE  ${\left\lbrace\mathbf{{X}_s^{l}}\right\rbrace}_{s=1}^{N_R}$ Relevant Subjects
    \STATE Estimate Sample Covariance Matrix from feature matrix as Eq.(\ref{eq:SCM})
    \STATE Apply the transformation as $\Sigma_i ^{(j)} \Rightarrow (\bar{\Sigma}^{(j)})^{-\frac{1}{2}} \Sigma_i ^{(j)} (\bar{\Sigma}^{(j)})^{-\frac{1}{2}}$
    \STATE Calculate the mean for each subject as $\bar{\Sigma}_s = \argmin_{\mathbf{\Sigma} \in \mathcal{M} } 
    \sum_{i=1}^{n_s} 
    \delta_R^2 \left( \mathbf{\Sigma},\mathbf{\Sigma}_i \right)$
    \STATE Use Riemannian distance as similarity metric and define Affinity matrix $A_{ij}=exp(-\frac{\delta_R^2 (\bar{\mathbf{\Sigma}_i},\bar{\mathbf{\Sigma}_j})}{\sigma^2})$
    \STATE Define $D=\text{diag}(d_{11},d_{22},\cdots,d_{NN})$, where     $d_{ii}=\sum_{j=1}^{N} A_{ij}$
    \STATE 
     Determine the number of cluster $K$ by counting the eigenvalues of $\mathbf{A}'=D^{-\frac{1}{2}} A D^{-\frac{1}{2}}$ which are most close to 1. 
    \STATE Compute the $K$ eigenvectors $\{ \mathbf{u}_j\}_{j=1}^{K}$ of $A'$ associated with its $K$ largest eigenvalues, and form the matrix $\mathbf{U}=[\mathbf{u}_1,\mathbf{u}_2,\cdots,\mathbf{u}_K]$ by stacking the eigenvectors in columns
    \STATE Apply k-means to the rows of $\mathbf{U}$ to cluster the data into $K$ different groups.
    \STATE Assign the original point $\Sigma_i$ on Riemannian manifold to cluster $j$ if and only if row $i$ of the matrix $U$ was assigend to cluster $j$ 
  \end{algorithmic}
\end{algorithm}

    \section{Experiment}
    To demonstrate the merit of the proposed approach, we used a public EEG database, the PhysioNet CHB-MIT database. This database contains EEG data with 23 channels from 23 patients divided among 24 cases (one patient has 2 recordings, 1.5 years apart) \cite{shoeb2009application}	(www.physionet.org/physiobank/database/chbmit/). The goal in this database is to detect whether a 10 second segment of signal contains a seizure or not with high sensitivity and specificity and low false negative rate, as annotated in the database.	
    
    {At the first step of pre-processing, a 5$^{th}$-order Butterworth 0.5-30 Hz band-pass filter  was applied. Each recording was divided into 10 sec epochs and classified as either dominantly seizure or non-seizure (using expert labels). Then, the FFT coefficients were extracted in three standard bands: theta (4-7 Hz), alpha (8-13 Hz) and beta (13-30 Hz). With a bin size of 0.1 Hz, this resulted in 250 Fourier coefficients for each of the 23 channels. These coefficients were then concatenated and covariance matrices extracted. Then to increase the similarity of the data between subjects, each covariance matrix was transformed per equation (\ref{eq:Transformation}). The subsets of relevant subjects were then determined per Algorithm \ref{alg:algo1}. After subject selection, a SVM classifier was trained on labeled data from the subjects that were located in the same cluster and then tested on the withheld patient (i.e via a leave-one-subject-out cross validation (LOSO-CV) procedure). In order to use many popular and efficient classifiers, most of the literature focuses on mapping the covariance matrices into a tangent space of Riemannian manifolds to extend Euclidean-based algorithms to the Riemannian manifold of the SPD matrices \cite{multi, TransferLearningRiemannian}. The SVM classifier can be applied on the tangent space located at the geometric mean of the whole set of trials from relevant subjects to a given test subject as follows: 
    	$\mathbf{\Sigma}_{\mathfrak{G}} = \mathfrak{G}(\mathbf{\Sigma}_i, i = 1, \cdots, I)$. Each SCM, $\mathbf{\Sigma}_i$, is then mapped into this tangent space, to yield the set of $m = \frac{n(n + 1)}{2}$ dimensional vectors \cite{multi}:
    	\begin{equation}
    	\label{TangentSpace}
    	s_i={\mathbf{\Sigma}_{\mathfrak{G}}}^{-\frac{1}{2}} log_{\mathbf{\Sigma}_{\mathfrak{G}}}(\mathbf{\Sigma}_i) {\mathbf{\Sigma}_{\mathfrak{G}}}^{-\frac{1}{2}}
    	\end{equation} 
    	In the experiments detailed here, the LIBSVM toolbox \cite{LIBSVM} was used}. 
    
    {Table \ref{tab:CHBMIT} provides the per-patient (LOSO-CV) results and table \ref{tab:CBHMITall} summarizes the average results and compares them to the state-of-the-art. The methods proposed previously by Chen {\it et al.} \cite{chen2017high} and Thodoroff {\it et al.} \cite{thodoroff2016learning} are based on the wavelet transformation and deep learning, respectively. Table \ref{tab:CBHMITall} shows an increase over previous works in accuracy and specificity by 2-3\%. (Subject-specific works are not included in this comparison, since training and testing on the same subject is less useful and inflates statistics.)  We also note that we improve the false positive rate from 1.7/hour to 0.77/hour over Shoeb's original work \cite{shoeb2009application}. To the best of our knowledge, the method described in this article is the first work to propose a subject selection on a Riemannian manifold for unsupervised cross-subject seizure detection.}
    \begin{table}[h!]
    	\scriptsize
    	\begin{center}
    		\caption{Performance on CHB-MIT database}
    		\label{tab:CHBMIT}
    		\begin{tabular}{|c|c|c|c|c|}
    			\hline
    			\textbf{Subject ID} & \textbf{Accuracy (\%)} & \textbf{Sensitivity (\%)} & \textbf{False Positive rate (seizures/h)} & \textbf{Latency (sec)} \\
    			\hline
    			\hline
    			1 & 93.33& 98.28 & 0.194 & 5.10
    			\\ \hline 2 & 84.47 &  100.00 & 0.43 & 7.52 
    			\\ \hline 3& 93.51 & 100.00& 0.26 & 2.63
    			\\ \hline 4& 91.41 & 84.14& 0.22 & 7.42 
    			\\ \hline 5& 94.25 & 100.00& 0.34 & 4.46
    			\\ \hline 6& 82.79 & 46.63& 1.74 & 3.01 
    			\\ \hline 7& 86.56 & 98.22& 1.48 & 5.62 
    			\\ \hline 8& 88.89 & 100.00& 0.35 & 4.02
    			\\ \hline 9& 95.41 & 100.00& 1.28 & 8.23
    			\\ \hline 10& 92.82 & 100.00& 1.20 & 2.87 
    			\\ \hline 11& 94.39 & 85.16& 0.46 & 2.52 
    			\\ \hline 12& 84.13 & 62.40& 2.34 & 5.63
    			\\ \hline 13& 90.62 & 83.53& 2.86 & 8.12 
    			\\ \hline 14& 84.73 & 56.33& 0.55 & 3.78 
    			\\ \hline 15& 86.13 & 78.49& 0.24 & 5.85
    			\\ \hline 16& 86.40 & 58.42& 1.66 & 3.34
    			\\ \hline 17& 90.84 & 81.19& 0.82 & 6.21
    			\\ \hline 18& 85.40 & 97.93& 0.41 & 5.13
    			\\ \hline 19& 91.84 & 100.00& 0.21 & 9.89 
    			\\ \hline 20& 93.76 & 69.59& 0.58 & 2.84 
    			\\ \hline 21& 93.62 & 100.00& 0.46 & 2.78
    			\\ \hline 22& 87.28 & 100.00& 0.52 & 12.44
    			\\ \hline 23& 92.76 & 68.79& 0.14 & 1.36
    			\\ \hline 24 & 90.76 & 89.38& 0.10 & 5.01 
    			\\ \hline mean$\pm$(std) & 89.84 $\pm$ (3.90) & 85.77$\pm$(16.96)&  0.77$\pm$(0.75)& 5.24$\pm$(2.65) \\
    			\hline
    		\end{tabular}
    	\end{center}
    \end{table}
    \begin{table}[ht]
    	\scriptsize
    	\begin{center}
    		\caption{\scriptsize Performance comparison of works on CHB-MIT database}
    		\label{tab:CBHMITall}
    		\begin{tabular}{|c|c|c|c|c|}
    			\hline
    			\textbf{Method} & \textbf{Accuracy (\%)} & \textbf{Sensitivity (\%)} 
    			& 
    			\textbf{Specificity (\%)} 
    			\\
    			\hline
    			\hline 
    			Chen et al \cite{chen2017high}& 86.83 \% & 85.29 \% & 87.38 \%
    			\\ \hline 
    			Thodoroff et al \cite{thodoroff2016learning} & 84.18\% & 85.16 \% & 83.21 \% 
    			\\
    			\hline 
    			Proposed Method & \textbf{89.84} \% & \textbf{85.77} \% &  \textbf{89.64} \% \\
    			\hline
    			
    		\end{tabular}
    	\end{center}
    \end{table}
    \section{Conclusion}
    It is well-known that EEG signals are very specific to each subject. As a result, establishing a generic population model with high classification performance is extremely challenging, due to the inherent inter-subject variability.
    In this work, a novel subject selection approach for cross-subject brain signal classification was proposed and tested. Spectral clustering on a Riemannian manifold was applied in order to identify subsets of relevant subjects to create sub-models. By evaluating the proposed algorithm on the CHB-MIT scalp EEG database, we have shown that the our method can outperform previous published works on the same data (not focused on individually specific models). Future work will focus on evaluating the proposed method on other databases and addressing how the mismatch between the distributions of training and test subjects may be reduced.
    \subsubsection*{Acknowledgments}
    This research is supported in part by funding from the James S. McDonnell Foundation, Grant $\#220020484$ (http://www.jsmf.org), the Rett Syndrome Research Trust and Emory University.
    \bibliographystyle{unsrt}
    \bibliography{cite}
\end{document}